\documentclass{article}
\usepackage[utf8]{inputenc}
\usepackage{graphicx}

\title{The subset-matched Jaccard index for evaluation of Segmentation for Plant Images}
\author{Jonathan Bell, Hannah Dee\footnote{Corresponding author: hmd1@aber.ac.uk. Address: Computer Science, Llandinam Building, Aberystwyth University, Penglais, Aberystwyth SY23 3DB. The authors would like to thank Edel Sherratt for her helpful comments on an earlier draft.}}
\date{\today}

\begin{document}

\maketitle
\abstract{
We describe a new measure for the evaluation of region level
segmentation of objects, as applied to evaluating the accuracy of leaf-level segmentation of plant images. The proposed
approach enforces the rule that a region (e.g. a leaf) in either
the image being evaluated or the ground truth image evaluated
against can be mapped to no more than one region in the other
image. We call this measure the subset-matched Jaccard index.
}

\section{Introduction}
This report introduces an approach to the evaluation of plant
segmentation images against ground truth images. The approach is
intended to be suitable for segmentation methods which subdivide
a plant into leaves. It was developed to evaluate region (leaf)
based segmentations such as those used in the Leaf Segmentation Challenge 
\cite{Minervini2015c,Scharr2014a}. The aim of this document is to provide background, mathematical detail and motivation for the segmentation method we propose.

This document accompanies the release of an implementation 
together with a substantial dataset of top down visible light 
timelapse images of growing 
\emph{Arabidopsis thaliana} (Arabidopsis) plants.
Further details of this are in an appendix to this document.
This software reports the region (leaf) counts of
both images as well as measures of the degree to which corresponding 
regions in the test and ground truth images actually do coincide.

\section{Evaluation of leaf level segmentation} \label{sec:data:eval}
Other plant image databases have suggested approaches to evaluation of 
segmentation. Here we introduce our preferred approach to this,
concentrating on leaf-level segmentation.
\subsection{Requirements} \label{sec:data:eval:reqs}
Leaf segmentation is a multi-part segmentation in which we want to evaluate the separation of plant
pixels into individual leaves.
The following are desirable features of an approach to 
evaluation measures for multi-part object classification:-
\begin{enumerate}
    \item Background pixels are excluded to 
    avoid the results being swamped by the prevalence of background in
    many images, as is done in LSC approach.
    \item Only perfect agreement between test and ground truth 
    images should give a perfect score.
    \item The measure should be commutative, that is, the result should be the same whichever of a pair of images is 
    the ground truth. This will avoid either over- or 
    under-segmentation being unfairly punished relative to the other.
    \item The approach should be equivalent to an established measure where 
    images are binary. (e.g. the approach amounts to either the Dice 
    coefficient or Jaccard index).
    \item The approach should normalise over the size of the object but not the size 
    of regions. Some measures, e.g. the LSC approach, would be skewed by a 
    small region having a poor Dice score, even though few pixels
    are wrongly classified.
\end{enumerate}
As a supplementary concern, we suggest that having the approach amounting to 
Jaccard (rather than Dice) is preferable. This means that the measure amounts to the number of correctly 
classified object pixels divided by number of pixels classified as 
object in either image. This preference is partly as it seems to fit 
better with treating classifications as sets, and partly because 
expanding the approach to multiple correct classifications is more 
natural. We also have a preference for not allowing more 
than one class in one image to be classified with a class in the 
other. Current approaches (e.g. LSC) allow this, and thus could count scores where \emph{all} leaves match just one ground truth leaf. Again, there are possible arguments against our suggestion approach as our technique implies that classifying an object class against a previously used 
object class is no better than classifying it as background.

Put simply, our approach amounts to an indication of the degree of similarity
between two sets (like Dice or Jaccard) but extended to indicate
the degree of similarity between the marked subsets of each set.
This is where the sets are those pixels classified as object (plant)
in each image and the subsets are the pixels classified as some 
region of the object (leaf).

One limitation we accept is that an evaluation comparison should consider 
only one plant (object). If there are several objects in an image, the region labels
(e.g. colours) for one object might map to different region labels in its annotation than those of 
a different object. This could be avoided by having each object use its own set of labels.
It is suggested that if images with several plants are to be 
used, these are better divided into single plant cropped images (This is problematic when plants overlap, of course).

\subsection{Alternative approaches} \label{sec:data:eval:other}
The leaf segmentation challenge approach 
\cite{Minervini2015c,Scharr2014a} simply takes the mean of the 
best Dice coefficient found for each region (leaf).
This means it is possible for more than one region in one image to be 
classified against some region in the other and it is also the case 
that the best Dice results are not symmetric. 
That is, the results differ depending on which of a pair
of images is treated the ground truth image.
They work around this by running their best dice function twice, 
swapping the images and keeping the worse result. 
This they call ``symmetric best Dice''.

They also find the difference between numbers of leaves in each image
and the plant-from-background Dice. These three give the same
results as our proposed approach.

The MSU-PID dataset paper \cite{Cruz2015a} has four evaluation metrics. 
One of these is the symmetric best Dice score and they suggest using 
Scharr's implementation. 
The other three measurements are different from the LSC ones and are leaf tip based:-
\begin{itemize}
    \item Unmatched leaf rate - percentage of unmatched leaves with reference to the total number of labelled leaves.
    \item Landmark error - average tip bases errors smaller than some threshold to indicate leaf tip alignment error.
    \item Tracking consistency - percentage of frame by frame
    correspondent leaves whose tip error is less than a threshold to
    indicate tracking consistency.
\end{itemize}

\subsection{The proposed approach} \label{sec:data:eval:app}
If we consider the measurements as being the degree of similarity of the sets of pixels classified
as a region in the segmentation and in the annotation,
the Jaccard index $J$ for object (plant) classification is the size of the intersect between $S$,
the set of pixel locations classified as object in the segmented image, and $T$ the set of 
pixel locations classified as object in the ground truth image divided by the
size of the union of $S$ and $T$.
    \[
        J = \frac{\vert S \cap T \vert}{\vert S \cup T \vert}
    \]
For leaf level segmentation, we can treat $S$ and $T$ as being divided into subsets
where each pixel in the set is in one such subset (it belongs to exactly one leaf).
This involves defining a mapping between $T$ our ground truth set and $S$ our
segmentation set.

$$
m: T \rightarrow S \\
$$

To avoid counting matches twice we define $m$\ as the best possible greedy
assignment of leaf segmentations in our ground truth to those in the target image.
The aim of this assignment is to ensure each detected leaf is matched to the closest
ground truth leaf, but no leaf in either set is matched more than once. 

\begin{eqnarray*}
\forall t \in T, s \in S : m(t)=s \Rightarrow \\
\forall s' \in S : J (t,s') \leq J (t,s) \wedge \\ 
\forall t, t' \in T : m(rt) = m(t') \Leftrightarrow t=t'
\end{eqnarray*}

The intersections of these $m$\ assignments are then summed to give $I$, a per-plant measure. 

$$
I = \sum_{m}{I^{s}_{m} \cap I^{t}_{m}}
$$

\begin{figure}[ht]
    \centering
    \includegraphics[width=\textwidth,keepaspectratio=true]{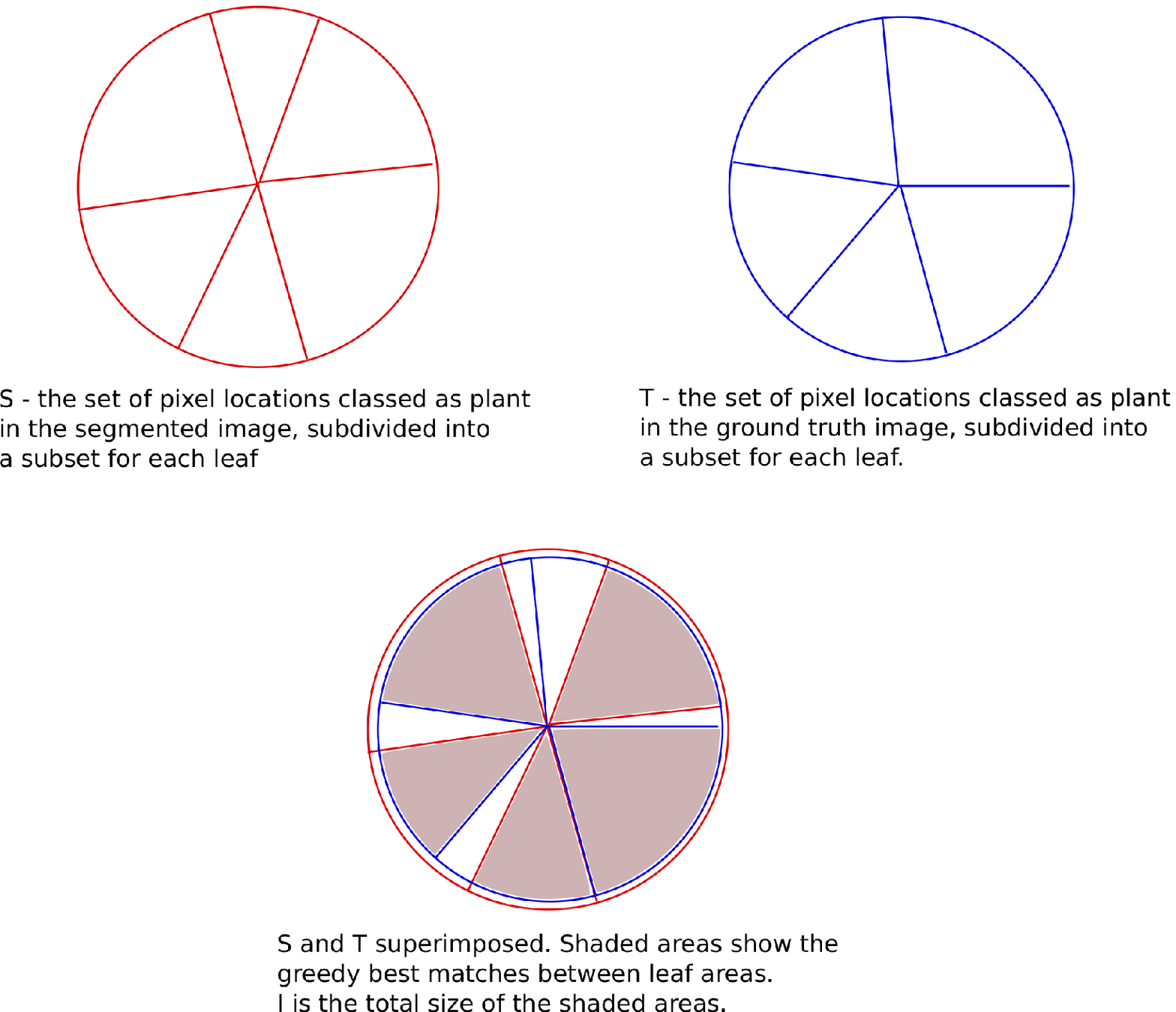}
    \caption{The sets of pixel locations in the segmented and ground truth images with
    how individual subset regions are matched to find the subset-matched Jaccard index.}
    \label{fig:subsets}
\end{figure}

For a plant level segmentation which takes into account the agreement of leaf detections, therefore, we define our measure, the subset-matched Jaccard index as 
    \[
        J_s = \frac{I}{\|S \cup T\|}
    \]
Figure \ref{fig:subsets} illustrates this approach.
The shaded areas are the regions where pixel location are common to matched subsets in the two sets of locations, so $I$ is the total area of these.
The region (leaf) subset at the top in the segmented set $S$ has no matches, as the only corresponding regions in the ground truth have been matched elsewhere.

This can also be visualised in terms of a confusion matrix. If we have a confusion matrix of leaf level classifications between the segmented and truth images, then $I$ is found by finding the highest total from this matrix using at most one value from each row and one value from each column. This is how the implementation works.
    
It is trivial to alter this to give similar subset-level Dice coefficients.
The object level Dice coefficient is
\[
    D = \frac{2\|S \cap T\|}{\|S\| + \|T\|}
\]
and the subset-matched Dice coefficient is
\[
    D_s = \frac{2I}{\|S\| + \|T\|}
\].

These values are obtained from a confusion matrix giving the size of each set of
segmented region's pixels classified as mapping to a given annotated region\footnote{In our implementation, the pixels of each region in an image share an integer value, these need not
be consecutive values.}.
The implementation has six steps, as follows:-
\begin{enumerate}
    \item Generate lists of $m$ colours used in the truth image
    and $n$ colours used in the evaluated image, with the 
    integer that represents black (it might not be zero) as
    the first item in each list. (This step is not needed with 
    consecutively numbered region classifications.)
    \item Generate a zero valued $m*n$ matrix $C$ to
    form a confusion matrix for pixel classifications. It has the 
    columns ordered by the order of colour values in the test image 
    list and the rows by the order in the truth image list. 
    This means $(1,1)$ will be for those pixels classified as 
    background in both images.
    \item Iterate over both images, identifying the pixel values of 
    each location in turn, finding the position of the respective 
    values in both lists and incrementing the corresponding 
    location in the matrix from step 2.
    \item The number of object regions (leaf count) in the segmented 
    image is $n-1$ and from the truth image is $m-1$.
    The absolute value of $m-n$ is the difference in number 
    of classifications (leaf counts) between the two images. If the 
    result is negative, the test image is over-segmented.
    \item To obtain the object level Jaccard index, the intersect and union
    of $S$ and $T$ can be obtained from the matrix $C$ as
    \[
        \|S \cap T\| = \sum\limits_{c=2}^m \sum\limits_{r=2}^n C_{c,r}
    \]
    and
    \[
        \|S \cup T\| = \sum\limits_{c=1}^m \sum\limits_{r=1}^n C_{c,r} - C_{1,1}
    \]
    Less formally, this is the sum of values from 2 to $n$ in 
    each row from 2 to $m$
    in $C$ divided by the sum of all values in $C$ except $(1,1)$.
    \item To get the subset-matched Jaccard index, 
    we find the maximum value for the total of the sizes of
    intersections between these sets of subsets. From $C$ this is
    the highest possible sum of values 2 to $n$ from each row 
    from 2 to $m$ but taking no more than one value from any row or 
    column. This is then divided by $\|S \cup T\|$ obtained
    as before.
\end{enumerate}
This implementation gives all three measurements (leaf count, object Jaccard and subset-matched Jaccard) from 2 iterations over each image,
so should be quicker than the LSC approach. The algorithm for
extracting the best total in step 6 is both recursive and iterative.
It is simpler than the Hungarian Method as there is no need to
identify which values are used, only the total is needed. There
is no ordering in this approach, but step 6 does ensure that no
region in either image can be mapped to more than one region in 
the other.

To give similar object subset-correlated Dice coefficients if preferred,

\begin{itemize}
    \item from $C$, $\|S \bigcap T\|$ is found as described and
    \[
        \|S\| = \sum\limits_{c=1}^m \sum\limits_{r=2}^n C_{c,r}
    \]
    and
    \[
        \|T\| = \sum\limits_{c=2}^m \sum\limits_{r=1}^n C_{c,r}
    \]
    \item We can define a subset-matched Dice coefficient $D_s$ analogously to the Jaccard one in step 6 above where $I_s$, $\|S\|$ and $\|T\|$ are obtained as described earlier.
\end{itemize}

\subsection{Testing and results}
The software has been tested upon pairs
of alternative ground truth images hand made by different people.
These were high throughput phenotype platform images with twenty
plants were image, so each image was cropped into twenty subsidiary
images - one for each plant. There is a tendency for pairs of images with a large
difference in numbers of regions found to score badly. This is to be expected and is indeed a feature of our approach.

In addition some test images were made to test and demonstrate 
that the expected results were achieved. These were:-
\begin{itemize}
    \item Evaluating two identical images (using the same image 
    twice) gives perfect results (Jaccard indices of 1) and
    difference between numbers of subsidiary regions is 0.
    \item Evaluating \texttt{image a} against \texttt{image b} gives the same 
    result as evaluating \texttt{image b} against \texttt{image a}. Except, of course, the 
    opposite image will have more segments so an over-segmentation
    will become an under-segmentation.
    \item If a segmented image is all background, it scores 
    zero against a non-blank truth image.
    \item If two all-background images are evaluated together, the 
    scores should be one. (This follows from 1 and is consistent with
    the definition of the Jaccard index.)
    \item An image segmented as all object will not score zero 
    against a truth image that is not all background.
\end{itemize}

\section{Discussion}
We believe our approach to be preferable to the LSC approach.
As already mentioned: the LSC approach is not symmetrical and so runs the comparison both ways and keeps the worse result. Our measure does not need this. 

We release software to implement these and other segmentation evaluations. Whilst there is no need to calculate both Jaccard and Dice scores (as
the two measures are functionally related \cite{PontTuset2015a}) we report both for ease of use. 

Thus the software supplied with our dataset returns the following results:-
\begin{itemize}
\item Region (leaf) count in test image.
\item Region (leaf) count in ground truth image.
\item Difference between region counts in the two images.
\item Object (plant) level Jaccard index.
\item Subset (leaf) level Jaccard index as described herein.
\item Object level Dice coefficient.
\item Subset level Dice coefficient.
\item LSC style ``symmetric best Dice'' score.
\end{itemize}

It is not feasible to use this approach to evaluate images with more
then one object (plant). This is because no ordering is assumed
so there is no reason to expect that a label will match the same 
label in a different plant. Since the approach relies on label,
this means several objects will tend to have a poor score. The same
broadly applies to the LSC approach. A suggested work-around
is to subdivide such an image (or, strictly, pair of images)
into single plant regions and take the mean of the index scores.
The number of subsidiary region's differences cannot really be
treated this way, though. An alternative would be to make sure each 
object (plant) has its own distinct set of labels. Even here
the possibility that some objects are over segmented while others are
under-segmented will mislead unless the absolute values are summed.

A possible approach to managing images with several objects
would be to extract regions of interest surrounding each
connected region of non-background values from the annotated image
and the corresponding regions from the image to be evaluated.
This assumes that all of an object's regions are connected in
the annotated image and that objects do not overlap.
The ground truth annotations for our dataset are not all connected
as in some cases the petioles (leaf stems) were partially buried
so this approach is not feasible for us.

The implementation includes a simple approach to dividing an image into a 
grid of regions according to the number of objects (plants) 
across and down the image.
Obviously the objects need to be in an evenly spaced grid.
It then returns two result files, a full set and a summary.
The full results are similar to those listed above for when a
segmentation of one object is evaluated with these additional columns:- 
\begin{itemize}
    \item Indication of where the two images agree on number of regions in 
    an image portion's object, marked with a 1.
    \item The amount by which the segmented image over segments. Only
    given a value where the test image does over-segment
    \item The amount by which the segmented image under segments. 
    Only given a value where the test image under-segments.
\end{itemize}
The full results include mean values and the count of 
objects where the numbers of regions in the two images regions agree
as a summary. Means of segmentation region count differences are taken 
from the objects with a result in the relevant column. So an
image of 20 plants of which three are over segmented by, say
2, 1 and 3 leaves will have a summary (mean) result of 2.
This means over and under segmentation of different objects do not cancel 
each other out.
There is also a separate summary result file that includes the number of 
objects that are over and under segmented as well as the summary results
similar to those in the full results file.


\section{Conclusion}
We believe that this proposed approach to region based segmentation evaluation is more rigorous than the LSC approach because each region in either image can be matched with at 
most one region in the other image. This also means results are 
symmetrical in the sense that the subset based Jaccard and Dice scores
are identical if the image treated as the test image is swapped with the 
ground truth image. Our matrix based implementation can also be used to obtain the LSC ``symmetric best Dice'' score and does so with fewer iterations over the image data.

\bibliographystyle{unsrt}
\bibliography{main}

\begin{thebibliography}{1}

\bibitem{Minervini2015c}
Massimo Minervini, Andreas Fischbach, Hanno Scharr, and Sotirios~A. Tsaftaris.
\newblock Finely-grained annotated datasets for image-based plant phenotyping.
\newblock {\em Pattern Recognition Letters}, 2015.

\bibitem{Scharr2014a}
Hanno Scharr, Massimo Minervini, Andreas Fischbach, and Sotirios~A. Tsaftaris.
\newblock Annotated image datasets of rosette plants.
\newblock Technical Report~-, Institute of Bio- and Geosciences: Plant Sciences
  (IBG-2), Forschungszentrum J{\"{u}}lich GmbH, J{\"{u}}lich, Germany, 2014.

\bibitem{Cruz2015a}
Jeffrey~A. Cruz, Xi~Yin, Xiaoming Liu, Saif~M. Imran, Daniel~D. Morris,
  David~M. Kramer, and Jin Chen.
\newblock Multi-modality imagery database for plant phenotyping.
\newblock {\em Machine Vision and Applications}, pages 1--15, 2015.

\bibitem{PontTuset2015a}
Jordi {Pont-Tuset} and Ferran Marques.
\newblock Supervised evaluation of image segmentation and object proposal
  techniques.
\newblock {\em {IEEE} transactions on pattern analysis and machine
  intelligence}, PP(99):1, 2015.
\newblock Pre-publication.

\bibitem{PSIweb}
{Photon Systems Instruments PlantScreen Phenotyping}.
\newblock http://psi.cz/products/plantscreen-phenotyping/ (accessed 13/4/2016).

\end{thebibliography}

\section*{Appendix: Arabidopsis plant image datasets}

This appendix provides overview information of image datasets which can be used to build leaf-level segmentation techniques.

\subsection*{Aberystwyth Leaf Evaluation Dataset}
As part of the work of carried out under grant number EP/LO17253/1 we have generated a dataset of several thousand top
down images of growing \emph{Arabidopsis thaliana} (Arabidopsis) plants
of accession Columbia (Col-0).
These images were obtained
using the Photon Systems Instruments (PSI) PlantScreen plant scanner 
\cite{PSIweb} at the National Plant Phenomics Centre situated on 
Aberystwyth University's Plas Gogerddan campus.

The plants were top view imaged using the visible spectrum every 15 minutes 
(nominally, in practice the interval was approximately 13 minutes)
during a 12 hour period. 
There are some gaps in the image sequence attributable to machine malfunctions.
The imaging resulted in the acquisition of 1676 images of each of 4 trays,
each image having 20, 18, 16, 14, 12 or 10 plants as plants were harvested.
Images were taken using the PSI platform's built in camera,
an IDS uEye 5480SE, resolution 2560*1920 (5Mpx) fitted with a
Tamron 8mm f1.4 lens. This set up exhibits some barrel distortion.
Images were saved as .png files (so using lossless compression)
with filenames that incorporate tray number (031, 032, 033 and 034)
and date and time of capture. Times are slightly different between 
trays as images were taken sequentially. Other than png compression, no
post-processing was done. This means images have the camera's barrel 
distortion. Code to correct this is supplied along with the dataset.

\begin{figure}[ht]
    \centering
    \includegraphics[width=\textwidth,keepaspectratio=true]{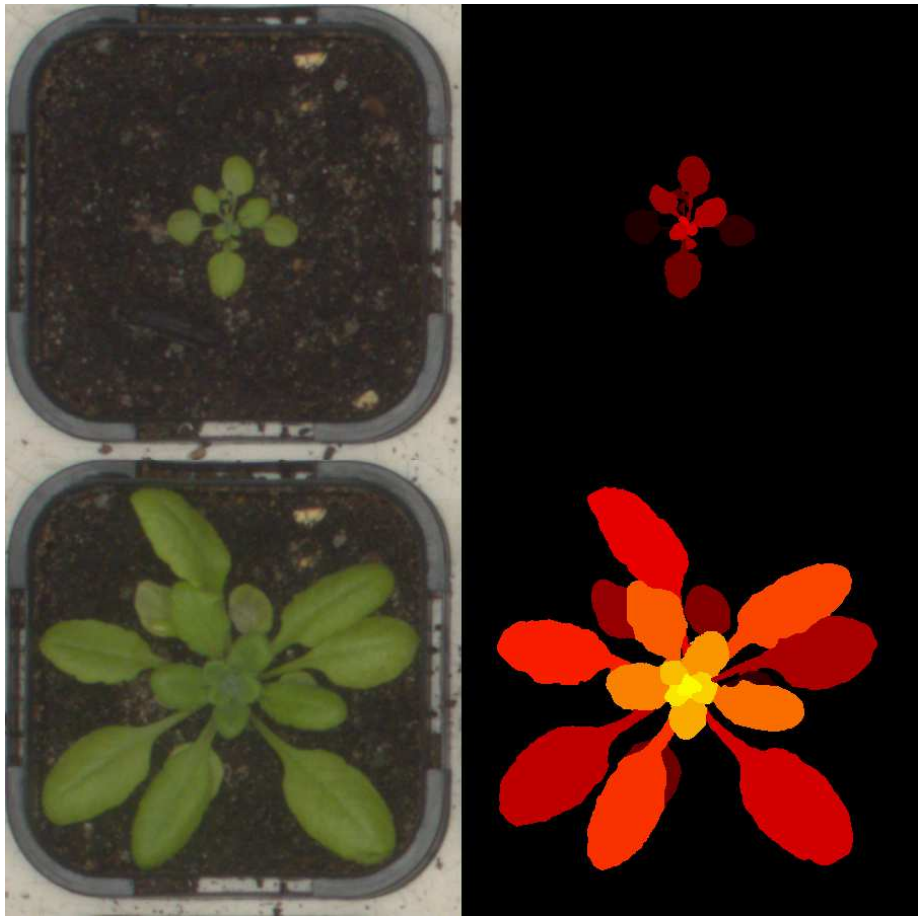}
    \caption{Two views of the same plant taken 27 and 42 days after sowing together with ground truth annotations.}
    \label{fig:ground_truths}
\end{figure}

Our dataset has accompanying ground truth annotations of the last image of 
each day of one tray (number 31), together with the first image taken 
after 2pm every third day. The suggestion is that these are used as 
training data. We also have ground truth annotations of the first image 
taken after 2pm every second day of tray 32 so these are usable as a test 
set, having no plants in common with the training data. 
This split means our dataset has 706 training data plant images 
available and 210 test data images.
It has been 
pointed out that the difference in times the annotated images
from the two trays were taken might result in a lack of correspondence 
caused by diurnal changes in leaf orientation (``hyponasty'').
Examination of the images suggests this is not the case, 
but an alternative approach would be to divide the images in to training 
and test portions. If this was done by halving the images, the test set 
size would be increased at the expense of the training data.
Examples of early and late growth images of a plant cropped from
the Aberystwyth Leaf Evaluation Dataset with annotations are shown in Figure 
\ref{fig:ground_truths}.

There is a fuller description of the dataset in its accompanying documentation.

\subsection*{Other Arabidopsis image datasets}
Two other datasets of Arabidopsis have been made publicly available.
In the datasets released for the leaf segmentation challenge (LSC)
\cite{Scharr2014a, Minervini2015c} were 161
top down visible light images of one set of Arabidopsis plants, 
40 of another set with different backgrounds and 83 images of tobacco.
These were split by the organisers into training and test sets
and the annotation data was only released for the training sets.
The splits were as follows:- A1, training 128, testing 33; A2
training 31, testing 9 and tobacco training 27, testing 56.
This means, of course, that all participants had exactly the same data.
The leaf segmentation challenge did not involve leaf tracking,
so time lapse sets of images were not used.
Although the images were taken using timelapse, 
no timelapse sequences have been released. 
Not all the data has been 
released to retain unseen data for the challenge itself. 
Images were chosen to exhibit features that present challenges to 
segmentation, such as moss on the soil.

The MSU-PID dataset \cite{Cruz2015a} includes multi-modality images of
Arabidopsis and bean plants. Beside visible light, they include
chlorophyll fluorescence, infra red and depth camera images.
Images were taken hourly throughout a 16 hour day.
The data is divided into a 40/60 split for training and testing. 
Specifically, 6 of the 16 Arabidopsis plants were
earmarked for training and 2 of the 5 bean plants. They use the
same evaluation metrics as LSC.

\end{document}